\documentclass[a4paper, 10pt, conference]{ieeeconf}      
\usepackage{FG2026}
\usepackage{multirow}
\usepackage{subcaption}

\usepackage{booktabs}      
\usepackage{multirow}      
\usepackage{graphicx}      
\usepackage{adjustbox}     
\usepackage{xcolor}        
\usepackage{amssymb}       
\usepackage{amsmath}       
\usepackage{ifthen}        
\usepackage{textcomp}
\usepackage{booktabs}
\usepackage{siunitx}
\usepackage[table]{xcolor}
\usepackage{caption}
\usepackage[normalem]{ulem}

\definecolor{bestcolor}{HTML}{006D6F}      
\definecolor{highlightcolor}{HTML}{F0F8FF} 
\definecolor{rulecolor}{HTML}{ADD8E6}       
\usepackage{booktabs}
\usepackage{siunitx}
\usepackage[table]{xcolor}
\usepackage{caption}
\usepackage[normalem]{ulem}

\definecolor{bestgreen}{HTML}{008000}       
\definecolor{secondbestgray}{HTML}{666666}  
\definecolor{highlightgray}{HTML}{F5F5F5}   

\definecolor{perfGood}{rgb}{0.0, 0.6, 0.0} 
\definecolor{perfBad}{rgb}{0.8, 0.0, 0.0}  

\newcommand{\arrow}[2]{{
  \ifthenelse{\equal{#1}{perfGood}}{\color{perfGood}}{\color{perfBad}}
  \ifthenelse{\equal{#2}{up}}{$\uparrow$}{$\downarrow$}
}}

\newcommand{\trendCCCGood}{\arrow{perfGood}{up}}   
\newcommand{\trendCCCBad}{\arrow{perfBad}{down}}   
\newcommand{\trendMAEGood}{\arrow{perfGood}{down}} 
\newcommand{\trendMAEBad}{\arrow{perfBad}{up}}     

\usepackage[skip=2.5pt]{caption}

\FGfinalcopy 

\IEEEoverridecommandlockouts                              
\def\FGPaperID{113} 

\title{\LARGE \bf
xTrace: A Facial Expressive Behaviour Analysis Tool for Continuous Affect Recognition
}

\author{\parbox{16cm}{\centering
    {\large Mani Kumar Tellamekala\footnote{Primary Author}$^{1*}$, Shashank Jaiswal$^1$, Thomas Smith$^1$, Timur Alamev$^1$, Gary McKeown$^2$, Anthony Brown$^1$, Michel Valstar$^1$}\\
    {\normalsize
    $^1$ Blueskeye AI, 
    $^2$ Queen's University Belfast, UK}}
}

\begin{document}

\ifFGfinal
\thispagestyle{empty}
\pagestyle{empty}
\else
\author{Anonymous FG2026 submission\\ Paper ID \FGPaperID \\}
\pagestyle{plain}
\fi
\maketitle

\begin{figure*}
    \centering
    \includegraphics[width=0.9\textwidth]{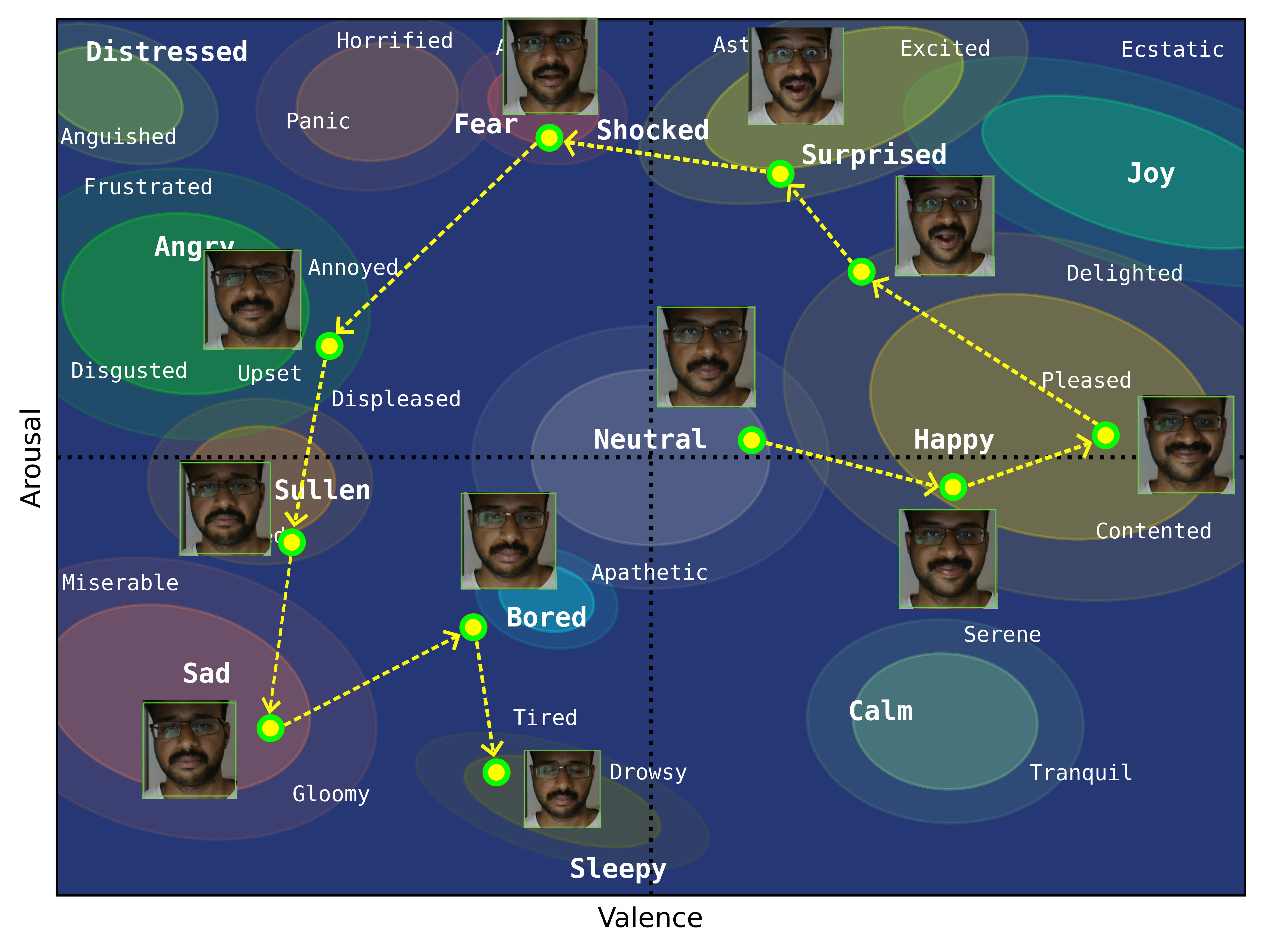}
    \captionof{figure}{Extensive predictive range of xTrace for continuous affect recognition within the 2D Emotion space}
    \label{fig:emo_zones}
  \label{fig:teaser}
\end{figure*}

\begin{abstract}
 Recognising expressive behaviours in face videos is a long-standing challenge in Affective Computing. Despite significant advancements in recent years, it still remains a challenge to build a robust and reliable system for naturalistic and in-the-wild facial expressive behaviour analysis in real time. This paper addresses two key challenges in building such a system: (1). The paucity of large-scale labelled facial affect video datasets with extensive coverage of the 2D emotion space, and (2). The difficulty of extracting facial video features that are discriminative, interpretable, robust, and computationally efficient. Toward addressing these challenges, this work introduces xTrace, a robust tool for facial expressive behaviour analysis and predicting continuous values of dimensional emotions, namely valence and arousal, from in-the-wild face videos. 

To address challenge (1), the proposed affect recognition model is trained on the largest facial affect video data set, containing $\sim$450k videos that cover most emotion zones in the dimensional emotion space, making xTrace highly versatile in analysing a wide spectrum of naturalistic expressive behaviours. To address challenge (2), xTrace uses facial affect descriptors that are not only explainable, but can also achieve a high degree of accuracy and robustness with low computational complexity. The key components of xTrace are benchmarked against three existing tools: MediaPipe, OpenFace, and Augsburg Affect Toolbox. On an in-the-wild benchmarking set composed of $\sim$50k videos, xTrace achieves 0.86 mean Concordance Correlation Coefficient (CCC) and on the SEWA test set it achieves 0.75 mean CCC, outperforming existing SOTA by $\sim$7.1\%. We present a detailed error analysis of affect predictions from xTrace, illustrating (a). its ability to predict emotions with high accuracy across most bins in the 2D emotion space, (b). its robustness to non-frontal head pose angles, and (c). a strong correlation between its uncertainty estimates and its accuracy. 
\end{abstract}

\section{Introduction}

\noindent Automatic recognition of facial expressive behaviour is a critical component in enabling empathetic Human-Computer Interactions. During the last three decades, several efforts have been made to build tools for recognising apparent affect or emotion from face videos~\cite{pantic2000automatic,valstar2006fully,sanchez2021affective,kollias2025advancements}. Despite the significant progress made so far, robust facial affect recognition in naturalistic conditions is still not a solved problem. Why is it challenging to build such a robust tool capable of handling real-world complexities? 

Affective states and their nuances encountered in everyday naturalistic interactions go beyond the seven basic emotions~\cite{ekman1999basic}: neutral, happy, angry, sad, disgust, fear, and surprise. Most of the existing methods in the literature~\cite{tian2011facial,jiang2020dfew,wang2022ferv39k} and the widely used tools for facial expression analysis (e.g. DeepFace\footnote{https://github.com/serengil/deepface}) largely focus on the seven basic emotions, severely limiting their applicability to real-world scenarios. Most importantly, due to its discrete nature, the basic emotion model fails to capture the richness of emotional expressions encountered in naturalistic interactions. The dimensional model of continuous emotions~\cite{russell1980circumplex} is considered one solution to address this limitation. As shown in Figure~\ref{fig:emo_zones}, the dimensional model has the ability to describe a wide range of emotions. It does so by representing affective states as points living in a two-dimensional Euclidean space constructed by the dimensions: Valence (positive vs. negative) and Arousal (calm vs. excited). Thus, the dimensional emotion model makes it possible to represent facial affect as continuous values in the 2D emotion space and time. 

The dimensional model has received a lot of interest in Affective Computing in recent years~\cite{nicolaou2011continuous,gunes2011emotion,toisoul2021estimation}, however, its application in real-world systems is still very limited. One main reason for this is the paucity of large-scale in-the-wild affect-labelled video datasets for facial expressive behaviour analysis. Existing video datasets are relatively small in scale, with limited coverage of affective states in the 2D emotion space. For example, the Aff-wild-2 dataset~\cite{kollias2018aff} has almost no data for the low-arousal quadrants (see Figure~\ref{fig:datasets_label_dists}). As a result, the models trained and validated on such datasets, despite showing impressive accuracy levels, offer limited practical use in recognising a wide range of apparent emotions. In this work, we introduce a novel facial expressive behaviour analysis tool, xTrace, a versatile tool for recognising expressed facial affect in real time. Towards addressing the aforementioned limitations of the existing video affect datasets, we use the largest in-the-wild facial affect dataset composed of $\sim$450k in-the-wild videos to train the continuous affect recognition model in xTrace. Thus, this model shows wide predictive range of affective states in the dimensional emotion space, as shown in Figure~\ref{fig:emo_zones}. 

From a system building point of view, to make xTrace robust, accurate, and reliable in real-world applications, we incorporate the following factors into its design: 

\begin{enumerate}
    \item  A wide predictive coverage in the 2D emotion space
    \item The use of interpretable affect features for explainability
    \item Robustness to partial facial occlusions induced by non-frontal head pose angles in unconstrained settings
    \item The ability to output predictive uncertainty estimates
    \item Invariance to image modality (RGB vs. Near-Infra Red)
\end{enumerate}

The affect recognition model in xTrace is extensively evaluated w.r.t. its input feature accuracy, its output accuracy variations across the emotion space with varying levels of granularity, its robustness to occlusions, and the quality of its predictive uncertainty estimates. The key components of xTrace, 2D face alignment, Action Unit (AU) recognition, and affect recognition, are extensively benchmarked against existing tools such as MediaPipe~\cite{lugaresi2019mediapipe}, OpenFace~\cite{openface} and Augsburg Affect Toolbox~\cite{mertes2024affecttoolbox}. The evaluation results show that xTrace outperforms these tools by significant margins on in-the-wild benchmarking datasets.  

xTrace is heavily optimised for on-device processing in real-time ($>=$30FPS) using only $\sim$433.5M FLOPS per frame, with cross-platform compatibility (x86 and ARM devices). Its executable runs on Windows, Mac and Linux, and it can be made available for research purposes through an academic license request process for responsible use.

In summary, the key merits of xTrace are as follows:

\begin{itemize}[noitemsep]

\item Demonstrates high prediction accuracy across all key affective states by achieving an average CCC of 0.86.

\item Achieves SOTA performance on the SEWA test set by outperforming the best existing method by $\sim$7.1\%.

\item Uses explainable and interpretable facial affect descriptors: uncertainty-aware landmarks and AU intensities.

\item Shows minimal accuracy degradation in the presence of facial occlusions induced by non-frontal head pose.

\item Outputs predictive uncertainty estimates so that downstream tasks can decide when to trust affect predictions.


\end{itemize}

\section{System Architecture}

\begin{figure*}
    \centering
    \includegraphics[width=0.9\linewidth]{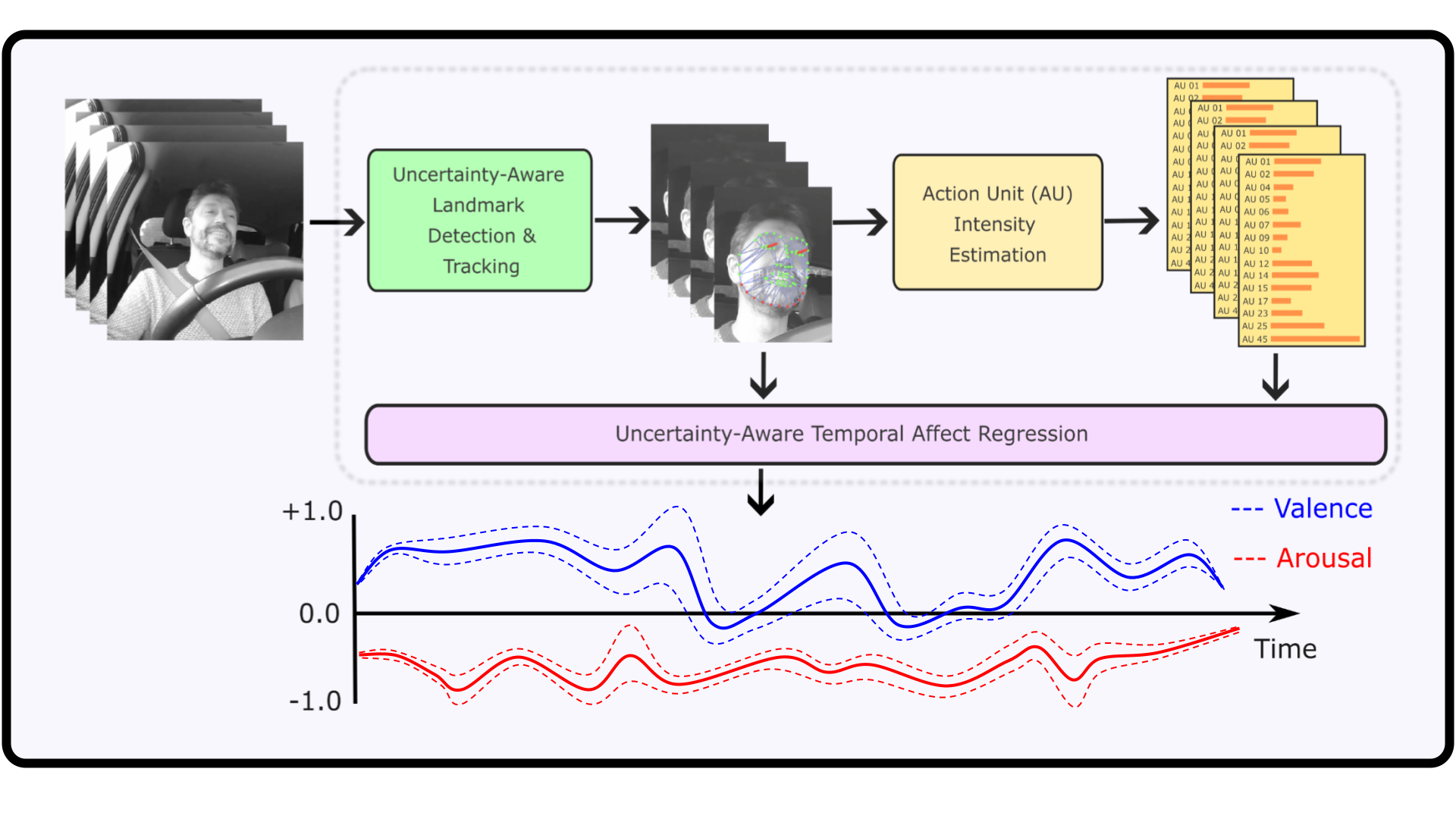}
    \caption{The end-to-end pipeline of xTrace for temporal valence and arousal estimation from face videos}
    \label{fig:sys_diag}
\end{figure*}

Taking a sequence of N consecutive video frames as input, xTrace outputs a sequence of VA predictions per frame and their corresponding uncertainty estimates. As illustrated in Figure~\ref{fig:sys_diag}, the end-to-end system architecture of xTrace is composed of three main modules: \textit{(1). Uncertainty-aware landmark regression (2). AU intensity estimation and (3). Temporal affect regression.} In this section, we first explain the rationale behind the key underlying design choices of xTrace and then describe each of these modules in detail.

\subsection{Key System Design Choices}

\noindent \textbf{Facial Affect Descriptors. }  Models trained to recognise affect from face videos use broadly two types of features extracted from images: Low-Level Descriptors (LLDs) and deep visual features. The former type of features are derived from spatio-temporal changes in 2D or 3D facial geometry and facial muscle actions. The latter set of features are the feature embeddings  directly extracted from cropped face images and/or image sequences using a deep neural network trained using large amounts of affect-labelled data. The commonly used low-level affect features in the literature are 68 2D facial landmark coordinates~\cite{jin2017face} and facial action unit intensity levels~\cite{ekman1977facial}. With the advent and success of deep representation learning, various deep learning approaches have been explored for facial affect recognition tasks~\cite{kollias2017recognition,mitenkova2019valence,tellamekala2019temporally,Kossaifi_2020_CVPR, toisoul2021estimation,kollias2025advancements}. Compared to deep feature embeddings, LLDs are more easily interpretable and grounded in known facial expressive cues. Also, as the dimensionality of LLDs is relatively low and they are not learned using emotion labels, they are more data-efficient than deep visual features. 

We chose to use LLDs as facial affect descriptors in xTrace, particularly 2D facial landmark features, their confidence scores, and AU intensities over deep visual features for the following reasons:  

\noindent \textbf{Explainability and Interpretability.} Unlike deep visual features, LLDs are easy to explain and interpret as they are standardised affect descriptors grounded in visible facial appearance changes. For instance, AU intensity features capture atomic movements of facial muscles and their temporal dynamics. When used as input features, AUs render transparency which deep visual features lack due to their black-box and hierarchically compositional nature. For example, humans easily interpret the meaning of specific facial actions such as a brow raise (AU1, AU2) or a smile (AU12), whereas deep visual features often lack such direct human interpretability. Given this advantage with LLDs, we can trace the affect recognition model's predictions back to its corresponding input features, and attribute them to visible appearance changes in different face regions. This traceability is particularly helpful in characterising the model's failure cases and improving the model's accuracy and robustness.

\noindent \textbf{Efficient use of labelled affect data.} When training an end-to-end model for affect recognition from face videos, the labelled data are used for supervision for two subtasks: learning per-frame visual features from face image pixels and learning temporal dynamics that are relevant to affect prediction. The former task involves distilling facial expression visual features that are robust to nuisance factors such as changes in head pose, lighting, occlusions, etc. To achieve such a high degree of robustness, it requires large amounts of affect-labelled data or sophisticated pre-training techniques (e.g., SwaV~\cite{caron2020unsupervised}) or joint-learning methods (e.g., EmoNet~\cite{toisoul2021estimation}). To achieve high affect recognition accuracy, it also requires large amounts of temporal variations in affect labels in training sets, to learn temporal dynamics of the facial features. Achieving both these objectives requires enormously large amounts of affect-labelled data, obtaining which is prohibitively expensive due to the requirement of trained human annotators for this task. To efficiently use the labelled data, in this work we chose LLDs whose models are trained to be robust to changes in head-pose angles, lighting, and occlusions. Thus, in the proposed system all the affect-labelled data gets mainly used for learning the temporal dynamics of affect labels.


\noindent \textbf{Agnostic to low-level image characteristics.} This is a key requirement to make an affect recognition system invariant to changes in the image acquisition process. For example, an end-to-end model trained on RGB images is likely to fail when applied to Near-Infrared (NIR) images of arbitrary wavelengths. This could be due to the dependence of deep visual features on low-level pixel statistics. Due to this limitation, an additional model fine-tuning effort may be required when there is a change in the image acquisition steps. In xTrace, by making the models that extract LLDs robust to such low-level pixel patterns, the affect recognition model is made agnostic to any low-level image characteristics.


\subsection{Affect Regression Outputs}

\noindent \textbf{Time- and value-continuous predictions of affect.} The affect regression model in xTrace is a sequence-to-sequence mapping model that takes a sequence of LLDs per frame as input and outputs a sequence of valence and arousal estimates. By mapping each frame to its corresponding continuous-valued vector in the 2D emotion space, this fine-grained approach allows us to make full use of the emotion representation capacity of the dimensional affect model. Also, by outputting per-frame affect vectors, this model captures transient and sudden changes in affective states over time. This level of spatio-temporal granularity broadens the applicability of xTrace for a broad spectrum of behavioural analysis tasks. 

\noindent \textbf{Sampling-free predictive uncertainty estimates of affect.} In addition to the point-wise affect predictions, xTrace also outputs three types of predictive uncertainty estimates of affect: Epistemic, Aleatoric, and cumulative~\cite{hullermeier2021aleatoric}. Most importantly, unlike other existing methods~\cite{kendall2017uncertainties,tellamekala2022dimensional}, it does so in a sampling-free manner, i.e., without requiring multiple forward passes through the model at inference time. This ability is important in obtaining uncertainty estimates with limited computational resources at inference time. Predictive uncertainties are essential for downstream applications to decide when they can or cannot trust the affect predictions. The three types of affect uncertainty output from xTrace have specific use cases. Epistemic uncertainty or model uncertainty allows the system to flag and collect out-of-distribution inputs and guide training data collection in later model update iterations. This feature is critical to continuously improve the coverage, accuracy, and robustness of the system over time. The aleatoric or data uncertainty captures the stochastic noise inherent in the input video. For example, when there is strong illumination on the face making it difficult to read the expression, the aleatoric uncertainty is expected to be high. When implemented in real-world applications, having access to this component of uncertainty can guide any potential automated interventions to improve the quality of face video. The cumulative uncertainty captures both epistemic and aleatoric components, and it can be made available to a downstream task to decide how reliable the affect predictions are for a given input.

\subsection{Building Blocks of xTrace}

\noindent \textbf{Uncertainty-Aware Landmark Detection and Tracking. } This module takes a video frame as input and outputs (a). a binary label for face shape validity (b). a 2D array of 68 landmark coordinates in iBUG format~\cite{sagonas2016300} and (c). a 1D array of landmark-wise uncertainty estimates. This module uses an edge-friendly CNN architecture with three output heads to jointly predict these three outputs. This CNN model is trained to be robust to in-the-wild facial appearances, on an internally curated face image dataset containing $\sim$650k RGB and NIR images labelled with 68 landmarks and their visibility scores. Note that the face-shape validity output is learned in an unsupervised fashion, whereas the landmark-wise uncertainty values are learned using explicit supervision. If the predicted face shape validity label is false for the current frame in the buffer, its corresponding affect features (both landmarks and AUs) are set to zeros and the next video frame in the buffer gets processed. If not, the predicted landmarks are used to preprocess the input to the AU intensity estimation module. It first extracts the face crop from the video frame using the bounding box computed from the previous frame's landmark coordinates. In case there is no previous frame or its predicted face shape validity is False, the bounding box is obtained from a face detection model heavily optimised for low inference latency and high detection accuracy on the edge devices.  

\noindent \textbf{AU Intensity Estimation } model takes an image of similarity-transformed face crop as input and outputs a 15-dimensional AU intensity vector. The predicted AU intensity outputs are continuous values in the range [0,5]; 0 indicates no activation and 5 indicates maximum activation. Similar to the landmark detection and tracking model, this module also uses an edge-friendly CNN architecture for compute-efficient model inference. By training on an internally curated  dataset that contains $\sim$2.4 million in-the-wild facial appearances, the AU model is made robust to the presence of partial occlusions and non-uniform illumination.

\noindent \textbf{Uncertainty-Aware Temporal Affect Regression} receives a sequence of temporal features, and outputs a sequence of VA predictions and their corresponding estimates of the epistemic, aleatoric, and cumulative uncertainty components. VA predictions are continuous values in the range [-1.0, 1.0], and their corresponding uncertainty estimates are clamped to the range [0, 1.0]. The input features to this model are prepared by normalising per-frame landmark coordinates and their corresponding uncertainty scores, and normalised AU intensity values. These normalised per-frame features are filled into a temporal buffer in a sliding window manner. 

\section{Video Datasets for Facial Affect Recognition}

\begin{figure*}
    \centering
    \includegraphics[width=0.9\linewidth]{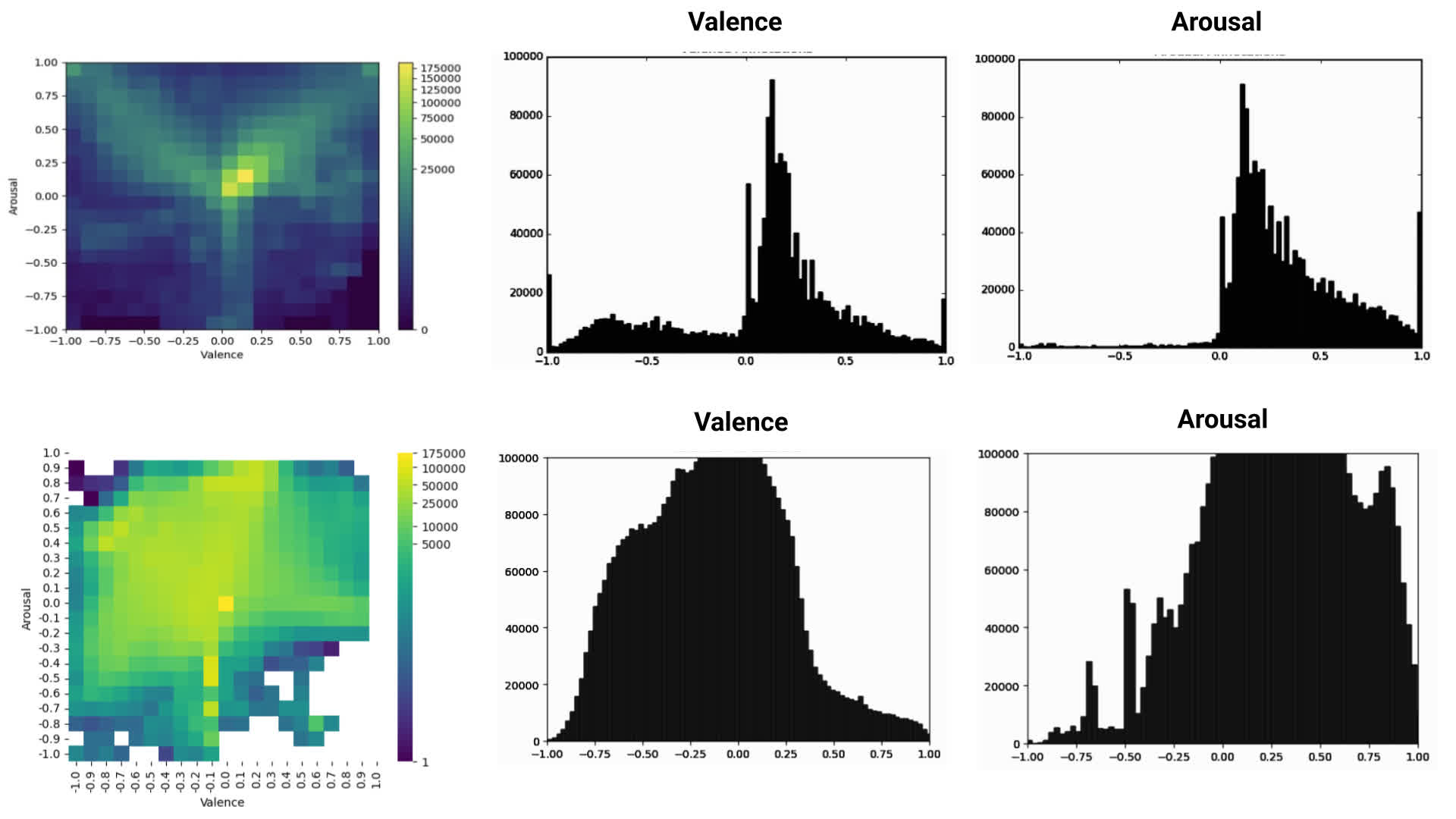}
    \caption{Comparison of emotion label distributions between the Aff-Wild2 dataset~\cite{kollias2025advancements}(top) and the xTrace dataset (bottom).}
    \label{fig:datasets_label_dists}
\end{figure*}



\begin{figure}
    \centering
    \includegraphics[width=0.8
    \linewidth]{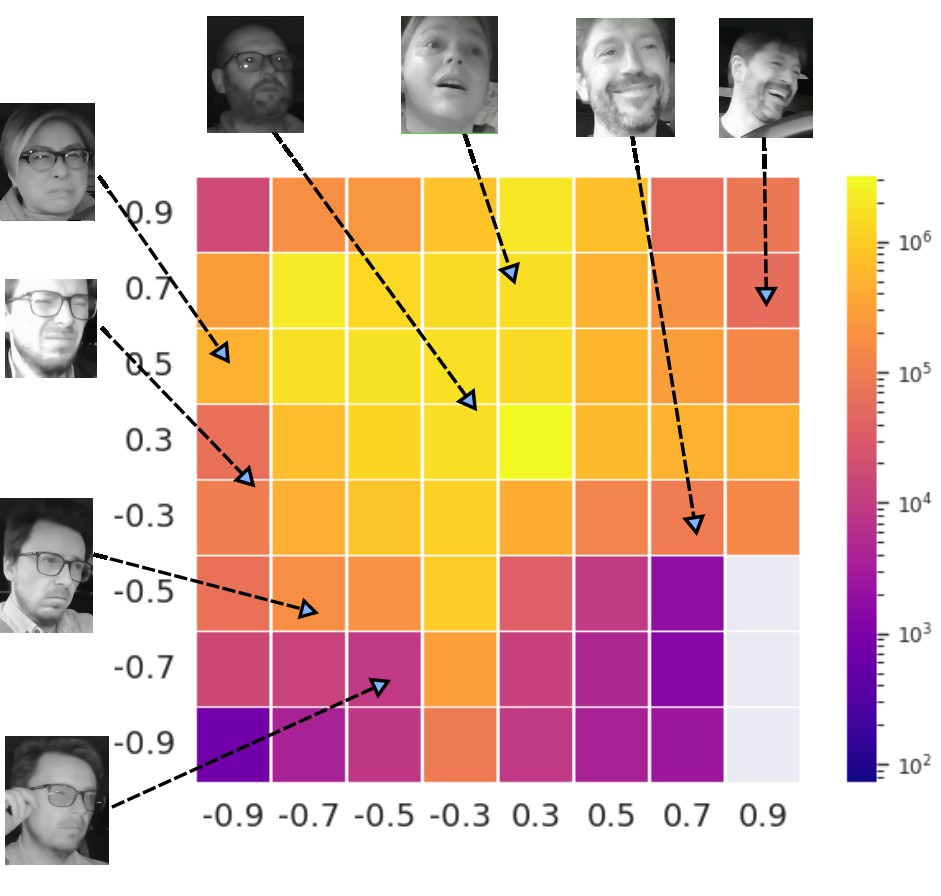}
    \caption{VA label distribution in the xTrace training dataset}
    \label{fig:train_set_dist}
\end{figure}

\noindent \textbf{Limitations of Existing Datasets.} SEWA~\cite{kossaifi2019sewa} and Aff-wild-2~\cite{kollias2018aff} are two prominent in-the-wild video datasets that are used to train and evaluate most existing facial affect recognition models. These datasets enabled numerous advances in affect recognition~\cite{kollias2025advancements}, especially the Aff-wild-2 dataset through a series of ABAW challenges~\cite{kollias2020analysing,kollias2021analysing}. However, they are still relatively small-scale datasets: SEWA contains 538 videos collected from 398 subjects, and Aff-wild-2 is composed of 594 videos from 584 subjects. Considering the complexity of naturalistic affect recognition as a learning task, such small-scale datasets may not be sufficient to cover the variations encountered in naturalistic and in-the-wild emotional expressions. Another key limitation of these datasets is their limited coverage of the valence and arousal space. As shown in Figure~\ref{fig:datasets_label_dists}, the label distributions of Aff-wild-2 clearly show insufficient coverage of the affect space. To make xTrace robust to real-world variations and cover the 2D emotion space as extensively as possible, we used a well-curated large-scale video dataset constructed from various in-house collected and curated face videos. 

\noindent \textbf{Expressive Video Clips Curation.} A set of $\sim$450k short video clips is curated from various in-house collected and in-the-wild face video datasets. These datasets were sourced from $\sim$8.5k subjects. The duration of the extracted short video clips varies in the range of 3 to 5 seconds. This video set also contains a significant proportion of NIR face videos captured in in-lab and driving (in-car) settings. To make sure that the low-arousal quadrants in the emotion space have a representative number of examples in this dataset, video clips sourced from an internal collected drowsiness dataset are also included. This dataset is partitioned into model development and benchmarking sets in a subject-independent fashion, resulting in $\sim$400k clips for model building and $\sim$50k clips for evaluation purposes. As a large proportion of this data set is composed of in-house data collected from external participants, For a variety of reasons, most notably GDPR, we are not able to make this dataset publicly available. We made sure that all the video data are sourced ethically from in-house data collection activities and other public sources with the required permissions.

It is worth notingh that Aff-wild-2 is available to only the ABAW challenge participants. Also, as Figure~\ref{fig:train_set_dist} shows, the label distribution of Aff-wild-2 is highly biased towards neutral emotion. Hence, we used the in-house constructed benchmarking dataset with $\sim$50k videos for thoroughly evaluating xTrace. Also, we used the SEWA test set for benchmarking xTrace against existing baselines in the literature.

\noindent \textbf{VA Annotation.} Considering the prohibitively expensive nature of trace-based per-frame VA annotation approaches, we employed a semi-automated annotation methodology to obtain per-clip VA emotion labels in a scalable approach. In this method, we used a simple web interface tool for single-point VA annotation in which each video clip is labelled with a point in the 2D emotion space. Compared to the label distributions of the Aff-wild-2 datasets, as shown in Figure~\ref{fig:datasets_label_dists}, the overall label distribution of this dataset has much better coverage in the VA space. Representative frames sampled from some example training video clips with different VA values are illustrated in Figure~\ref{fig:train_set_dist}.

\noindent \textbf{Inter-Rater Disagreement Analysis.} To evaluate the quality of VA labels obtained using the single-point annotation method described above, we performed an inter-rater disagreement analysis. For this purpose, we extracted a subset of $\sim$30k video clips annotated by at least three human raters. Since the labels annotated here are not traces but single points, instead of the standard correlation measures, we adopted a simple weighted mean absolute error (WMAE) method for measuring the inter-rater disagreement scores. Here, the weight values used in the MAE function are derived from the inverted distance values between the corresponding points. Given three VA annotations for a video clip, we computed WMAE scores using the weights derived inter-rater reliability scores~\cite{shrout79} calculated for each annotator. On this $\sim$30k video subset, the average WMAE values obtained for valence and arousal dimensions are 0.17 and 0.19 respectively. These low WMAE values indicate good inter-rater agreement levels of the annoated VA labels.





\section{Evaluation Results}

\subsection{Evaluation of Facial Affect Descriptors}

\begin{figure}
    \centering
    \includegraphics[width=1.0\linewidth]{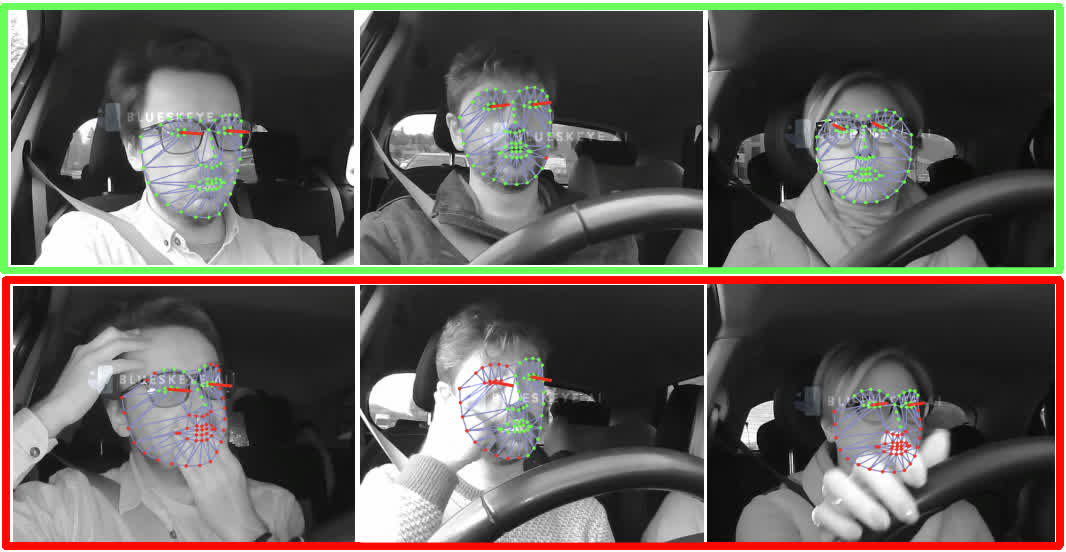}
    \caption{Landmark uncertainty estimates quality in xTrace}
    \label{fig:landmark_uncert}
\end{figure}

\noindent \textbf{Uncertainty-Aware Landmark Detection Evaluation } results are presented in Table~\ref{tab:face_tracker_eval_results}. For this evaluation, we used an internally curated benchmarking dataset composed of $\sim$39k in-the-wild RGB and NIR images. Both metrics, Area Under Curve (AUC) and Normalised Mean Error (NME), clearly show that the landmark detection model in xTrace achieves noticeably better point localisation accuracy despite its low computational complexity. Figure~\ref{fig:landmark_uncert} illustrates the ability of landmark-wise uncertainty estimates in detecting occluded facial regions, by comparing visible and occluded landmarks.

\begin{table}[htbp]
    \centering

    \renewcommand{\arraystretch}{1.3} 
    \sisetup{detect-weight, mode=text, text-series-to-math=true}
    
    \begin{tabular}{l S[table-format=1.2] S[table-format=2.2] S[table-format=1.2] S[table-format=2.2]}
        \toprule
        & \multicolumn{2}{c}{\textbf{RGB Test Set}} & \multicolumn{2}{c}{\textbf{NIR Test Set}} \\
        \cmidrule(lr){2-3} \cmidrule(lr){4-5}
        \textbf{Model} & {\textbf{NME$\downarrow$}} & {\textbf{AUC$\uparrow$}} & {\textbf{NME$\downarrow$}} & {\textbf{AUC$\uparrow$}} \\
        \arrayrulecolor{rulecolor}\midrule\arrayrulecolor{black}
        OpenFace 2.2.0~\cite{openface} & 7.03 & 55.46 & {4.56} & {60.29} \\
        MediaPipe~\cite{lugaresi2019mediapipe} & {4.36} & {55.77} & 6.62 & 46.69 \\
        \midrule 
        \rowcolor{highlightcolor}
        \textbf{xTrace (Ours)} & {\color{black}\bfseries 1.59} & {\color{black}\bfseries 88.72} & {\color{black}\bfseries 1.12} & {\color{black}\bfseries 89.68} \\
        \bottomrule
    \end{tabular}
    \caption{Comparison of facial landmark detection performance. Metrics: Normalised Mean Error (NME) $\downarrow$ - lower is better, and Area Under the Curve (AUC) $\uparrow$ - higher is better.}
    \label{tab:face_tracker_eval_results}
\end{table}

\begin{figure}
\centering
\begin{subfigure}[b]{0.49\textwidth}
   \centering
   \includegraphics[width=0.9\linewidth]{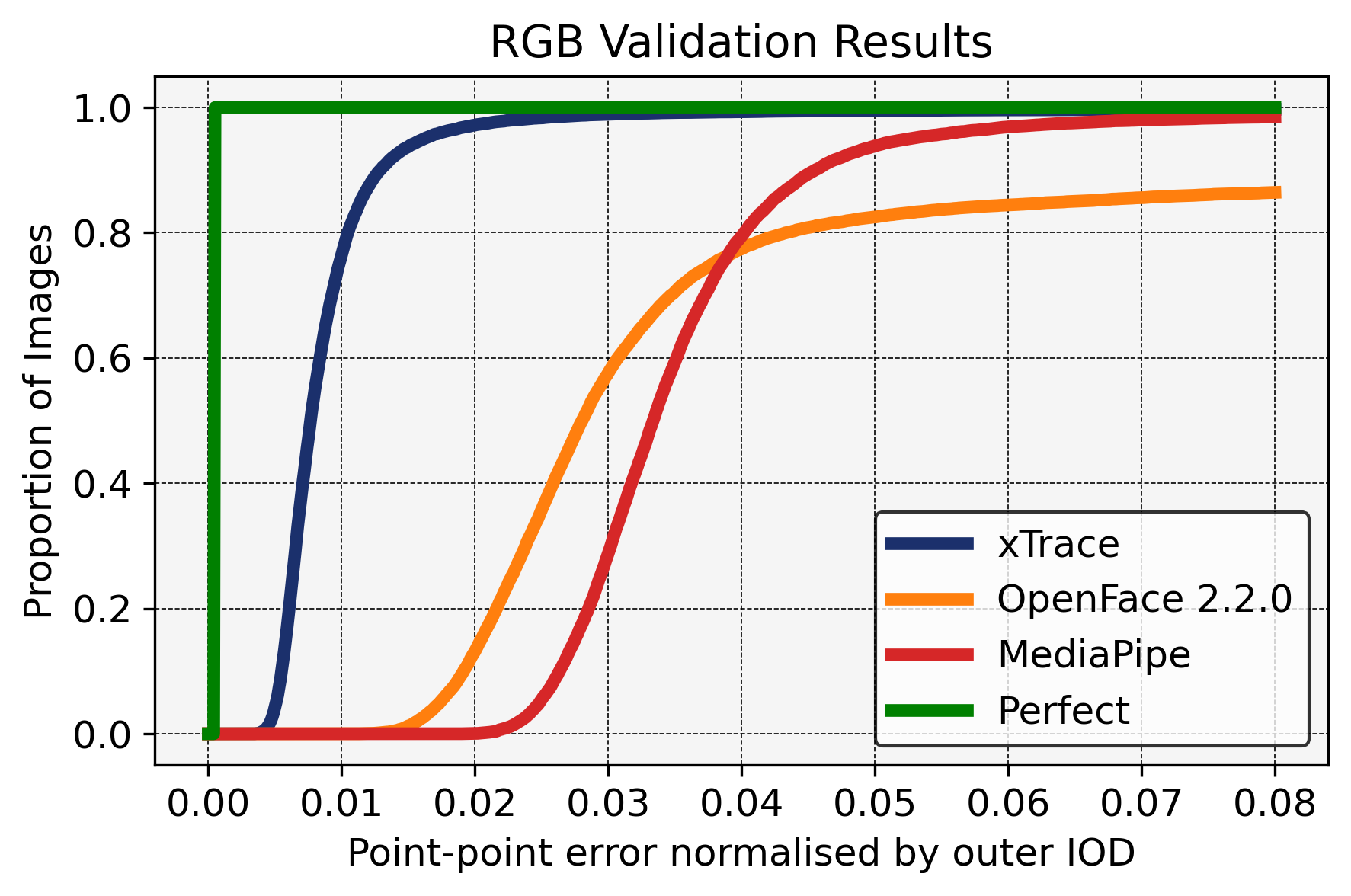}
   \caption{}
\end{subfigure}

\begin{subfigure}[b]{0.49\textwidth}
   
   \centering
   \includegraphics[width=0.9\linewidth]{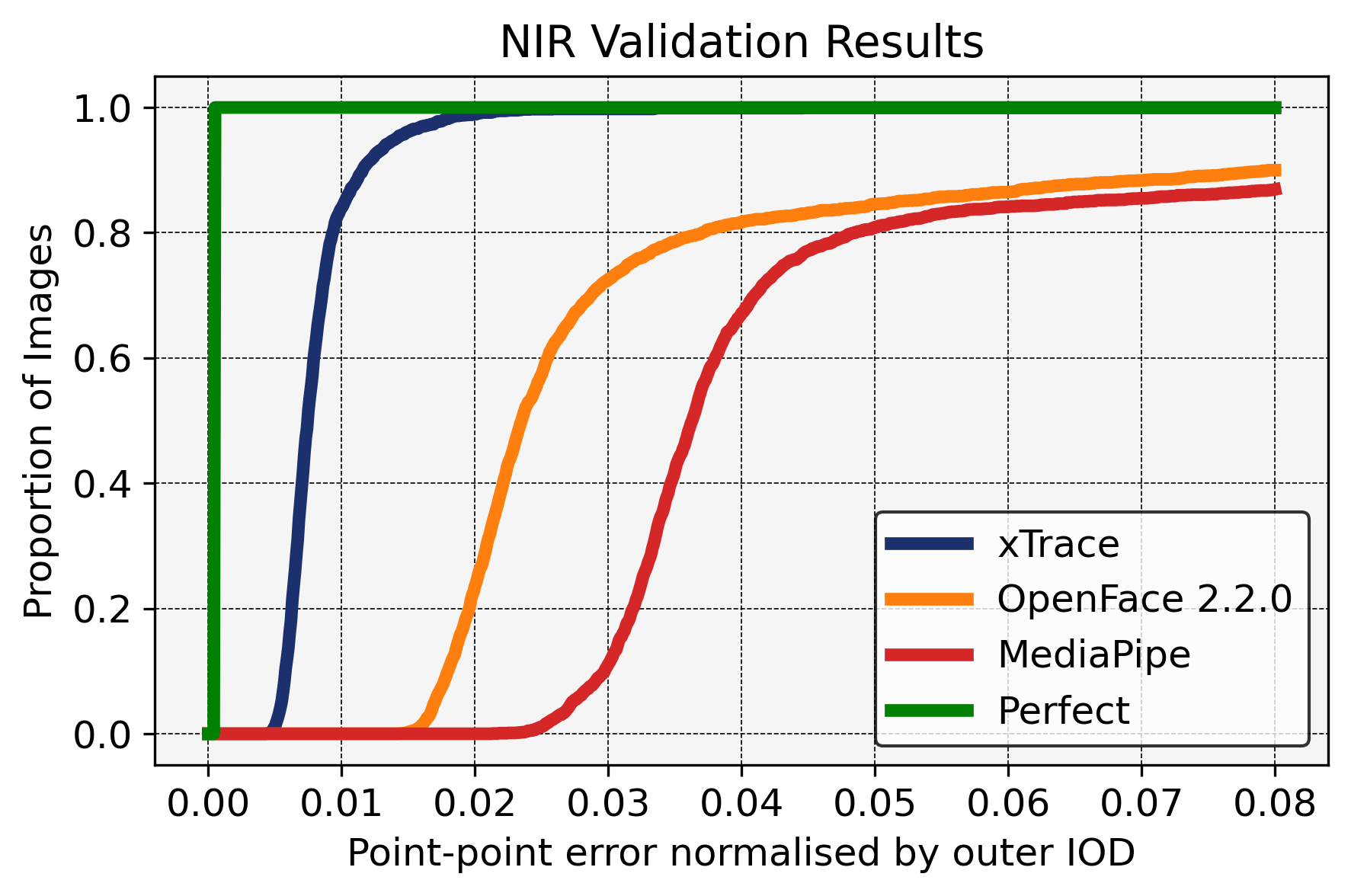}
   \caption{}
\end{subfigure}

\caption[Two numerical solutions]{Cumulative Error Distribution (CED) plots of landmark detection models on (a) RGB and (b) NIR images.}
\label{fig:ced_plots}

\end{figure}

\noindent \textbf{AU Intensity Estimation Evaluation } results are shown in Table~\ref{tab:au_res}, in terms of AU-wise and mean Intra-Class Correlation (ICC) scores. This in-house evaluation set contains $\sim$175k RGB and NIR face images annotated with AU intensity values of 15 AUs. Except for two AUs (07 and 14), for all other AUs the ICC scores are above 0.5. Overall, these high ICC scores show the strong predictive accuracy of the AU intensity model used in xTrace, and it outperformed OpenFace by $\sim$27\% in terms of the mean ICC score. 

\begin{table*}[!htbp]
    \centering
    \renewcommand{\arraystretch}{1.3}   
    \sisetup{detect-weight, mode=text} 
    
    \begin{tabular}{l *{16}{S[table-format=1.2]}}
    \toprule
    \textbf{AU $\rightarrow$} & {\textbf{01}} & {\textbf{02}} & {\textbf{04}} & {\textbf{05}} & {\textbf{06}} & {\textbf{07}} & {\textbf{09}} & {\textbf{10}} & {\textbf{12}} & {\textbf{14}} & {\textbf{15}} & {\textbf{17}} & {\textbf{23}} & {\textbf{25}} & {\textbf{45}} & {\textbf{Avg.}} \\
    \arrayrulecolor{rulecolor}\midrule\arrayrulecolor{black}
    \textbf{OpenFace 2.2.0} & 0.35 & 0.27 & 0.38 & 0.42 & 0.74 & 0.30 & 0.25 & 0.22 & 0.71 & 0.16 & 0.57 & 0.56 & 0.13 & 0.59 & 0.46 & 0.41 \\
    \rowcolor{highlightcolor}
    \textbf{xTrace (Ours) } & \bfseries 0.76 & \bfseries 0.70 & \bfseries 0.69 & \bfseries 0.66 & \bfseries 0.75 & \bfseries 0.47 & \bfseries 0.68 & \bfseries 0.53 & \bfseries 0.79 & \bfseries 0.42 & \bfseries 0.76 & \bfseries 0.54 & \bfseries 0.93 & \bfseries 0.78 & \bfseries 0.79 & \bfseries 0.68 \\
    \bottomrule
    \end{tabular}
    \caption{Evaluation results of the AU intensity estimation model in xTrace and comparison with OpenFace 2.2.0}
    \label{tab:au_res}
\end{table*}

\begin{figure}
    \centering
    \includegraphics[width=1.0\linewidth]{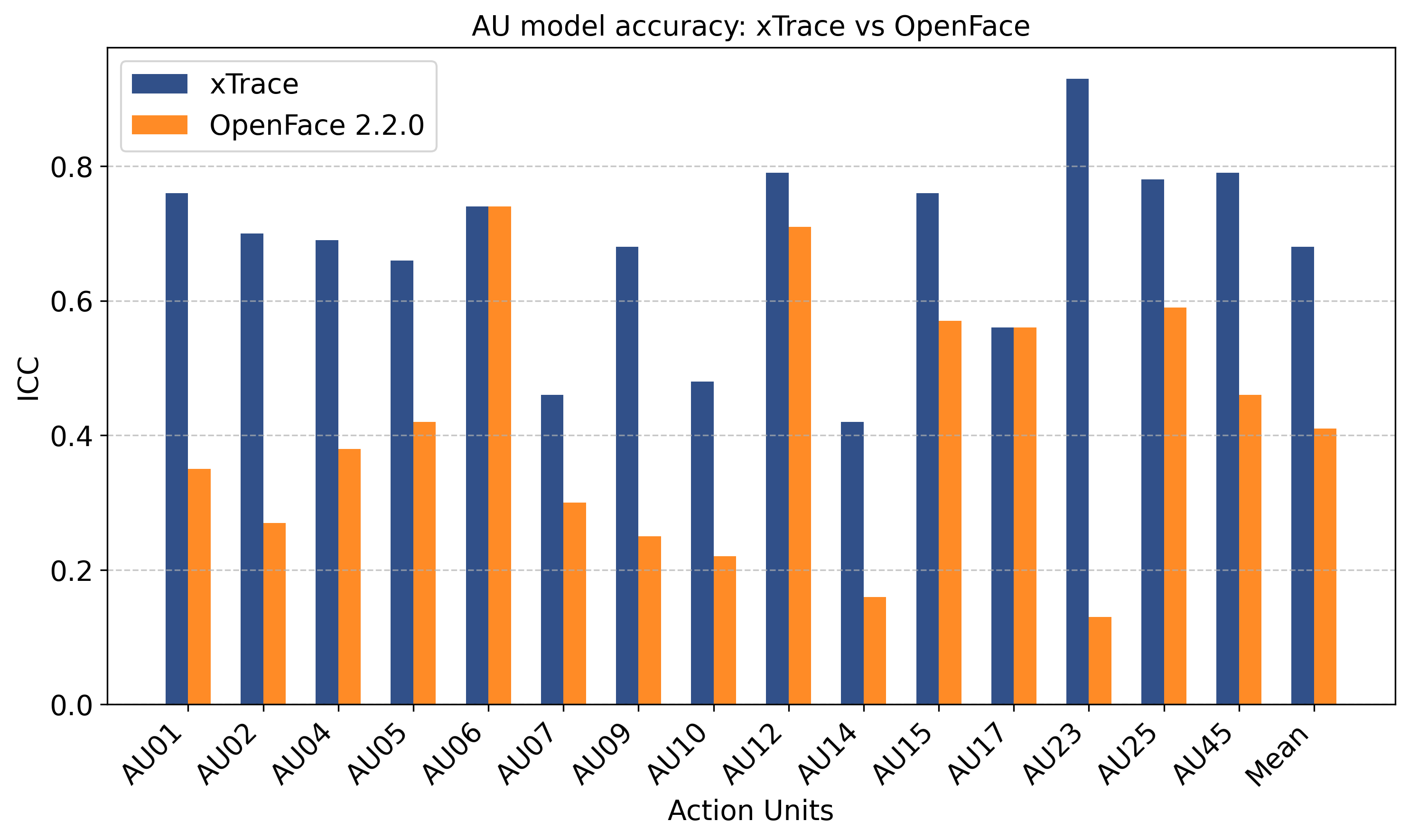}
    \caption{AU model accuracy comparison with OpenFace}
    \label{fig:au_lidar_openface}
\end{figure}

\subsection{Affect Recognition Evaluation}

\begin{table}[htbp]
    \centering
    \renewcommand{\arraystretch}{1.3}
    \sisetup{detect-weight, mode=text, text-series-to-math = true}

    \begin{tabular}{l S[table-format=1.2] S[table-format=1.2] S[table-format=1.2]}
        \toprule
        \textbf{Method} & {\textbf{Vale. CCC}} & {\textbf{Arou. CCC}} & {\textbf{Avg. CCC}} \\
        \midrule
        Augs. Aff. Toolbox~\cite{mertes2024affecttoolbox} & 0.62 & 0.11 & 0.37 \\
        Mitekova et al.~\cite{mitenkova2019valence}       & 0.44 & 0.39 & 0.42 \\
        Toisoul et al.~\cite{toisoul2021estimation}        & 0.65 & 0.61 & 0.63 \\
        Kossafi et al.~\cite{Kossaifi_2020_CVPR}        & 0.75 & 0.52 & 0.64 \\
        ResNet-Aff. Proc.~\cite{tellamekala2022modelling}     & 0.73 & 0.58 & 0.66 \\
        HGFAN-Aff. Proc.~\cite{tellamekala2022modelling}      & 0.74 & 0.63 & 0.69 \\
        Affective Processes~\cite{sanchez2021affective}   & 0.75 & 0.64 & 0.70 \\
        \midrule
        \rowcolor{highlightcolor}
        \textbf{xTrace (Ours)}  & {\color{black}\bfseries 0.81} & {\color{black}\bfseries 0.68} & {\color{black}\bfseries 0.75} \\
        \bottomrule
    \end{tabular}
    \caption{VA benchmarking results on the SEWA test set.}
    \label{tab:sewa_results}
\end{table}

\begin{figure}
    \centering
    \includegraphics[width=1.0\linewidth]{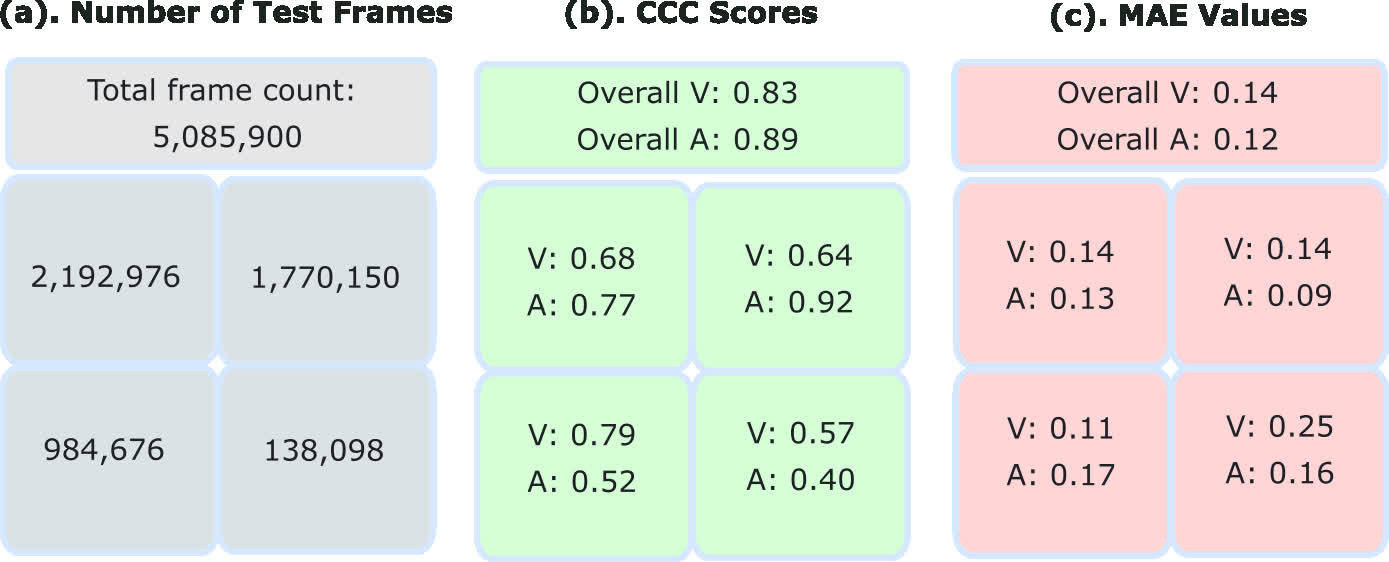}
    \caption{Emotion quadrant-wise VA evaluation results}
    \label{fig:quad_ccc_mae}
\end{figure}

\begin{figure}
\centering
\begin{subfigure}[b]{0.38\textwidth}
   \includegraphics[width=1\linewidth]{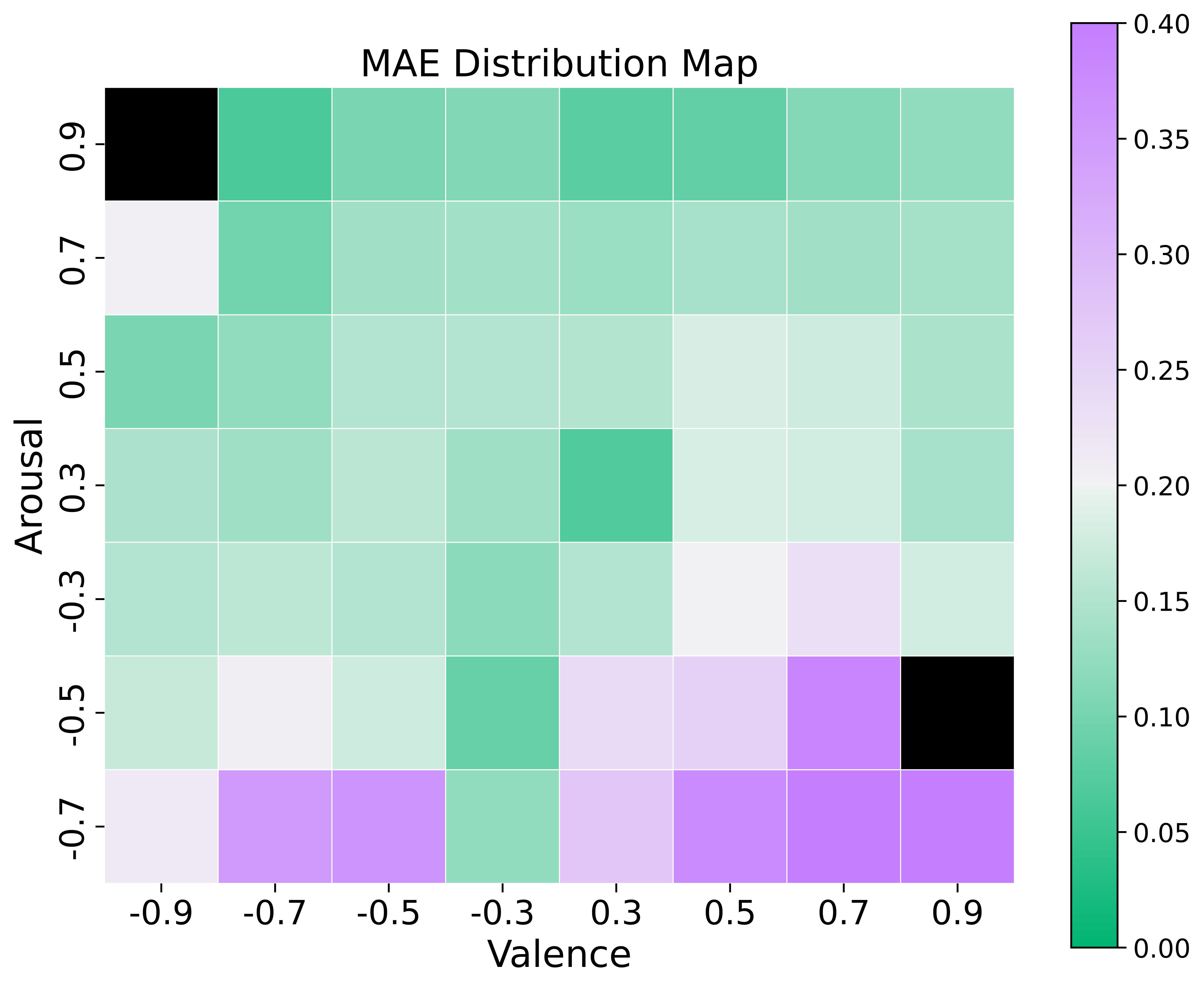}
   \caption{MAE distribution on the VA benchmarking set}
\end{subfigure}

\begin{subfigure}[b]{0.38\textwidth}
   \includegraphics[width=1\linewidth]{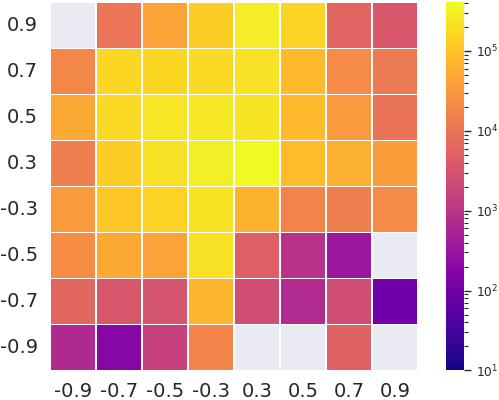}
   \caption{Sample distribution of the VA benchmarking set}
\end{subfigure}

\caption[]{Fine-grained VA evaluation results}
\label{fig:va_mae}
\end{figure}

\noindent \textbf{Overall VA Accuracy Evaluation Results } across the 2D affect space are presented in Figure~\ref{fig:quad_ccc_mae} in terms of two standard evaluation metrics~\cite{kollias2025advancements,toisoul2021estimation}: Concordance Correlation Coefficient (CCC) and Mean Absolute Error (MAE). While CCC captures the overall agreement between the model predictions and ground truth labels of all the test samples, MAE is computed between individual (per-frame) predictions and then averaged for all the samples. In total, this internally curated benchmarking set contains $\sim$5 million frames from $\sim$50k in-the-wild test videos. This benchmarking set is carefully constructed to cover as many bins as possible in the 2D affect space (see the sample distribution maps in Figure~\ref{fig:va_mae}.b.). The CCC scores and MAE values for both valence and arousal illustrate the strong predictive accuracy of the affect recognition model in xTrace.

\noindent \textbf{SEWA Benchmarking Results} are presented in Table~\ref{tab:sewa_results}. This test set contains 49342 frames labelled with valence and arousal, sourced from 53 in-the-wild videos of subjects from different cultures. This evaluation clearly shows that xTrace outperforms strong existing baselines such as Affective Processes~\cite{sanchez2021affective} by a large margin (7.1\% relative improvement in mean CCC). In addition, xTrace performs substantially better than the Augsburg Affect Toolbox, almost twofold. These results illustrate the superior performance of xTrace on a standard facial affect recognition benchmarking dataset.

\noindent \textbf{Quadrant-wise VA Evaluation Results} are presented in Figure~\ref{fig:quad_ccc_mae}: (a). number of test frames (b). CCC scores and (c). MAE values for each quadrant. It is important to note that this evaluation set contains significantly more frames in the positive arousal quadrants compared to the negative arousal quadrants. This imbalanced label distribution can also be seen in the training set (see Figure~\ref{fig:datasets_label_dists}).

The CCC scores and MAE values show the differences in the model accuracy across the four quadrants. For example, the valence CCC is the highest for the negative valence and negative arousal quadrant. However, the arousal CCC is the highest for the positive valence and positive arousal quadrant. In particular, in the case of arousal CCC the differences between the quadrants are more noticeable compared to the valence dimension. Interestingly, in the MAE values, the trends in cross-quadrant differences are a bit different compared to the CCC scores. These imbalances in model performance across different quadrants may be partly due to the imbalances in the training dataset. Also, it is worth noting that the visual modality (i.e., face videos) is known to be relatively less informative for arousal than valence. 


\begin{figure*}
    \centering
    \includegraphics[width=1.0\linewidth]{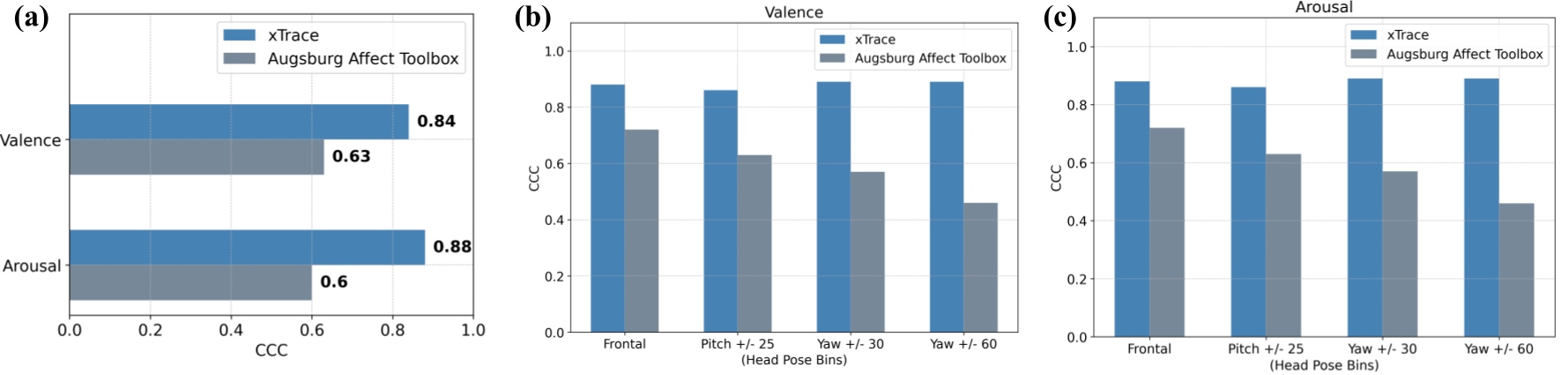}
    \caption{xTrace vs. Augsburg Affect Toolbox: (a). overall CCC and (b) and (c). robustness to non-frontal head pose angles.}
    \label{fig:lidar_vs_augsburg}
\end{figure*}

More fine-grained VA evaluation results (MAE values) are presented in Figure~\ref{fig:va_mae}, along with the sample distribution map of the evaluation set. Here, we used MAE scores to compare the model's accuracy across different VA bins. The bins with MAE values below and above human inter-rater disagreement (Weighted MAE) are marked in green and purple colors respectively. This result clearly illustrates that xTrace performs well in most VA bins, except for a few bins in the low-arousal quadrants. It could be due to the limitations of the visual modality in inferring arousal~\cite{pantic2000automatic}.

\begin{table}[]
    \centering
    \begingroup
    \renewcommand{\arraystretch}{1.3}    
    \begin{tabular}{l c c c c}
    \toprule
    & \multicolumn{2}{c}{\textbf{CCC}$\uparrow$} & \multicolumn{2}{c}{\textbf{MAE}$\downarrow$} \\
    \cmidrule(lr){2-3} \cmidrule(lr){4-5} 
    \textbf{Head Pose} & \textbf{Vale.} & \textbf{Arou.} & 
    \textbf{Vale.} & \textbf{Arou.} \\
    \hline
    Frontal &  0.86 & 0.86 & 0.16 & 0.15 \\
    Yaw +30 deg  & 0.85 & 0.85 & 0.18 & 0.15  \\
    Yaw -30 deg  & 0.88 & 0.82 & 0.15 & 0.16 \\
    Yaw +60 deg  & 0.84 & 0.84 & 0.19 & 0.15 \\
    Yaw -60 deg  & 0.87 & 0.78 & 0.16 & 0.18 \\
    Pitch +25 deg  & 0.86 & 0.84 & 0.17 & 0.15 \\
    Pitch -25 deg  & 0.83 & 0.84 & 0.18 & 0.16 \\
    \bottomrule
    \end{tabular}
    \caption{VA evaluation with varying head pose angles}
    \label{tab:head_pose_ccc_mae}
    \endgroup
\end{table}

\noindent \textbf{VA Accuracy Robustness to Non-Frontal Head Pose} is analysed in Table~\ref{tab:head_pose_ccc_mae} in terms of the CCC scores and the MAE values. The nonfrontal head pose angles considered in this evaluation cover the yaw bins [+60, +30, 0, +30, -60] and the pitch bins [+25, 0, -25]. For a fair comparison, it is ensured that this internal evaluation set is curated for each pose bin so that it has a roughly equal number of video clips ($\sim$3k clips per pose bin). Both CCC and MAE values show that the VA model's accuracy is highly robust to changes in the head pose angles. It is worth noting that even for the yaw +/- 60 degree range, the VA predictive accuracy drop is very low, compared to the frontal bin's VA accuracy. 

\noindent \textbf{Leave N-in Validation Results of VA Uncertainty} estimates are presented in Table~\ref{tab:uncert_eval}. In this evaluation protocol, only N\% of the test examples are considered for evaluation based on how high or low their cumulative uncertainty scores are. The CCC and MAE values that correspond to the `Lowest Uncertainty Filter' show how well the model performs when we consider only N\% of test examples with the lowest uncertainty. Similarly, the results of the `Highest Uncertainty Filter' show how worse the model performs when only the examples with the highest uncertainty are considered for evaluation. The trends in the CCC and MAE values in Table~\ref{tab:uncert_eval} for different N values show that the cumulative uncertainty scores are strongly correlated with accuracy. This strong correlation suggests that we can assess the reliability of xTrace's affect predictions in downstream applications.

\begin{table} 
\centering 
    \caption{Leave-N-in validation results of VA uncertainty}
    \label{tab:uncert_eval}
\setlength{\tabcolsep}{3pt} 
\small 
\begin{tabular}{@{}l l l c@{}} 
\toprule
 \textbf{Dimension} & \textbf{Metric} & \textbf{Filter Type}  & \textbf{Acc. Trend(N:25\%$\to$100\%)} \\
\midrule
\multirow{4}{*}{\shortstack{Valence}} 
                 & \multirow{2}{*}{CCC} & Lowest Unct.  & 0.93\,\trendCCCBad\,0.81 \\ 
                 &                      & Lowest Unct. & 0.61\,\trendCCCGood\,0.81 \\
                 \cmidrule(lr){2-4} 
                 & \multirow{2}{*}{MAE} & Lowest Unct.  & 0.08\,\trendMAEBad\,0.15 \\
                 &                      & Lowest Unct. & 0.21\,\trendMAEGood\,0.15 \\
\midrule
\multirow{4}{*}{\shortstack{Arousal}}
                 & \multirow{2}{*}{CCC} & Lowest Unct.  & 0.94\,\trendCCCBad\,0.88 \\
                 &                      & Lowest Unct. & 0.73\,\trendCCCGood\,0.88 \\
                 \cmidrule(lr){2-4} 
                 & \multirow{2}{*}{MAE} & Lowest Unct.  & 0.07\,\trendMAEBad\,0.14 \\
                 &                      & Lowest Unct. & 0.20\,\trendMAEGood\,0.14 \\
\bottomrule
\multicolumn{4}{@{}p{\dimexpr\linewidth-2\tabcolsep}}{\fontsize{7}{8}\selectfont {Accuracy Trend: \textcolor{perfGood}{Green $\uparrow \downarrow$} - Improving accuracy, \textcolor{perfBad}{Red $\uparrow \downarrow$} - Degrading accuracy}} \\
\end{tabular}

\end{table}


\section{Conclusion}

In this paper, we introduced xTrace, a robust, explainable and computationally lightweight tool for in-the-wild naturalistic facial expressive behaviour analysis. xTrace is trained on a corpus of $\sim$450k face videos, the largest video dataset for affect recognition to date. This makes xTrace capable of handling challenges encountered in real-world settings, ranging from subtle and nuanced expressive cues to partial occlusions caused by non-frontal head pose angles, etc. Some key system design choices we made in building xTrace include using Low-Level Descriptors of facial expressions (for model interpretability and compute efficiency) and outputting sampling-free uncertainty estimates (for reliability in downstream applications). Through extensive benchmarking of xTrace's components against existing tools such as MediaPipe, OpenFace and Augsburg Affect Toolbox, we demonstrated its superior accuracy in face alignment, AU intensity estimation and continuous affect recognition tasks. We believe xTrace will accelerate the adoption of facial affect recognition in a wide range of real-world applications supporting emphathetic human-computer interactions.

\noindent \textbf{Ethical Impact Statement.} xTrace has several real-world applications to support empathetic Human-Computer interactions. We commit to promoting ethical guidelines and responsible practices for the application of this tool. The intended use cases of xTrace include, but are not limited to, mobile applications for mental health and wellbeing, social robotics, driver monitoring systems, etc. It is heavily optimised for on-device video processing, which is essential to ensure data privacy in the aforementioned use cases.





{\small
\bibliographystyle{ieee}
\bibliography{egbib}
}

\end{document}